\pdfoutput=1

\documentclass[11pt]{article}

\usepackage[final]{acl}

\usepackage{times}
\usepackage{latexsym}
\usepackage{kotex}
\usepackage{todo}
\usepackage{xcolor}
\usepackage{xspace}
\usepackage{booktabs}
\usepackage{multirow}
\usepackage{url}
\usepackage{comment}

\usepackage[T1]{fontenc}

\usepackage[utf8]{inputenc}

\usepackage{microtype}

\usepackage{inconsolata}

\usepackage{graphicx}
\usepackage{tcolorbox}
\tcbuselibrary{skins, breakable}

\newif\ifcommentsoff
\commentsofftrue

\title{Culture is Everywhere: A Call for Intentionally Cultural Evaluation}

\author{
\begin{tabular}{c}
    Juhyun Oh$^\diamond$\quad Inha Cha$^\dagger$\quad Michael Saxon$^\ddagger$ \\
    Hyunseung Lim$^\diamond$\quad Shaily Bhatt$^\ast$\quad Alice Oh$^\diamond$
\end{tabular}
\\[2.5ex]
$^\diamond$KAIST \quad
$^\dagger$Georgia Institute of Technology \\
$^\ddagger$University of Washington \quad
$^\ast$Carnegie Mellon University
\\[1ex]
\texttt{411juhyun@kaist.ac.kr}
}

\begin{document}
\maketitle

\begin{abstract}
The prevailing ``trivia-centered paradigm'' for evaluating the cultural alignment of large language models (LLMs) is increasingly inadequate as these models become more advanced and widely deployed. Existing approaches typically reduce culture to static facts or values, testing models via multiple-choice or short-answer questions that treat culture as isolated trivia. Such methods neglect the pluralistic and interactive realities of culture, and overlook how cultural assumptions permeate even ostensibly ``neutral'' evaluation settings.
In this position paper, we argue for \textbf{intentionally cultural evaluation}: an approach that systematically examines the cultural assumptions embedded in all aspects of evaluation, not just in explicitly cultural tasks.
We systematically characterize the what, how, and circumstances by which culturally contingent considerations arise in evaluation, and emphasize the importance of researcher positionality for fostering inclusive, culturally aligned NLP research.
Finally, we discuss implications and future directions for moving beyond current benchmarking practices, discovering important applications that we don't yet know exist, and involving communities in evaluation design through HCI-inspired participatory methodologies.
\end{abstract}

\section{Introduction}
Language model-based applications are growing in adoption across the world. To ensure they are adopted responsibly and effectively, an understanding of their cultural impacts and sensitivities is important. Cultural misalignments in AI can perpetuate stereotypes, marginalize underrepresented voices, and fail to address the needs of diverse user communities~\cite{blodgett-etal-2020-language}. In response, the NLP and ML communities have begun to focus on culturally-aligned NLP, a subfield that aims to develop and evaluate systems capable of understanding and appropriately applying cultural knowledge in context \cite{adilazuarda2024towards, liu2024culturally, zhou2025culturetriviasocioculturaltheory}. The overarching goal is to create NLP systems that can effectively respond to and operate within varied cultural settings~\cite{bhatt-diaz-2024-extrinsic}. 
In this paper, we concentrate specifically on evaluation, as it increasingly shapes the direction of LLM development and deployment across diverse cultural contexts.

A key challenge, however, is that any decision in the evaluation pipeline—no matter how technical or routine—can carry cultural assumptions or consequences. 
For example, the tasks selected for evaluation often reflect the developers’ cultural context, which may not align with the needs of users from different backgrounds~\cite{hershcovich2022challenges}. Metrics assumed to be universal, such as what counts as ``well-structured'' writing, can vary significantly across cultures. Even expectations around interaction style and communication can differ~\cite{folk2025cultural,ge2024culture}, affecting how users perceive model outputs.

Despite this, the community often overlooks these \textit{cultural contingencies}, focusing attention only on the most obvious or explicit cultural questions (those labeled as ``cultural tasks'' or ``multilingual settings''). 
As a result, most current evaluation practices reduce culture to static facts, trivia, or proxies like nationality—primarily testing models through isolated factual questions~\cite{zhou2025culturetriviasocioculturaltheory} or their performance on culturally-cued prompts~\cite{mukherjee2024cultural}.
While knowledge of cultural facts is important, it fails to recognize the cultural contingencies embedded in seemingly ``neutral'' evaluation choices.

In this position paper, we argue that \textbf{every evaluative choice
should be examined for culturally contingent considerations}, not just those in explicitly cultural domains. We argue for a shift toward \textbf{\textit{intentionally cultural evaluation}}: a systematic approach that foregrounds cultural context throughout the evaluation process. By this, we mean making the cultural context of every evaluative decision explicit and deliberate, rather than leaving cultural influences implicit or accidental.

To challenge the current focus on only the most obvious choices like explicitly cultural tasks or multilingual settings, we systematically distinguish and discuss three key aspects of evaluation: (1) \textbf{what} is evaluated (\autoref{sec:what}), (2) \textbf{how} it is evaluated (\autoref{sec:how}), and (3) \textbf{in what circumstances} (\autoref{sec:where}) the desideratum is defined. We also examine the critical role of \textbf{researcher positionality} in shaping these evaluative choices (\autoref{sec:situated}). Finally, we outline the broader implications of our proposed approach for NLP research and practice (\autoref{sec:implication}). 

\paragraph{Contributions.}
Our work offers several key contributions.
We characterize the cultural contingencies in evaluation and propose building blocks for intentionally cultural evaluation. 
We find most evaluations reflect a narrow set of cultural assumptions, shaped by those who define the tasks and metrics. 
The design of ``what'' gets evaluated is frequently informed by dominant Anglocentric perspectives, reifying specific knowledge types and communicative norms while marginalizing others. We show that standard computational practices, such as static reference examples or aggregate metrics, are poorly equipped to assess culturally grounded variation, and argue for reimagining these methods to support more flexible, context-sensitive judgments of model quality. 
Crucially, we argue that culture in evaluation is not merely static content to be measured but is fundamentally tied to the circumstances of evaluation. We show how culture is both embedded in the very language of evaluation and enacted through culturally-contingent interactional patterns. As such, evaluating only static outputs misses key aspects of cultural alignment.

Finally, we call for greater reflection on the positionality of those evaluating. Evaluation of cultural competence in NLP is not neutral---it is shaped by the positionality of researchers and by systemic biases embedded in the broader AI/ML ecosystem. Researchers from lower-resource or non-Anglophone contexts often face pressure to conform to English-centric benchmarks to gain visibility, placing additional burdens on their work and constraining the development of research agendas grounded in local cultural contexts. This marginalization limits the diversity of perspectives represented in NLP and reinforces existing inequities.

Further, we suggest implications for moving beyond decontextualized methodologies toward more situated and culturally responsive methods, surfacing ``unknown unknowns,'' and co-constructing evaluation practices with affected communities. We ground our suggestions using findings from HCI studies. In doing so, we support a broader shift in NLP evaluation toward `thick evaluation'~\cite{qadri2025case}---an approach that prioritizes context-sensitive, community-aligned assessments of AI systems. 


\section{What to evaluate}\label{sec:what}
To move toward culturally intentional evaluation, we must ask: \textit{What tasks contain important, culturally contingent considerations?}
Current evaluations suffer from (a) overly narrow conceptions of `cultural' tasks and (b) externally imposed definitions of relevance, thus failing to capture true cultural competence in real-world contexts.

\subsection{Narrow Definitions of ``Cultural Tasks''}\label{sec:narrow_def}
Current evaluation practices suffer from a flawed dichotomy. On one hand, explicitly ``cultural'' tasks are often reduced to testing factual knowledge, a ``culture as trivia'' approach~\cite{zhou2025culturetriviasocioculturaltheory} that neglects the complex interaction patterns and behavioral expectations core to cultural competence. 
On the other hand, widely-used benchmarks such as MMLU \cite{hendrycks2021measuringmassivemultitasklanguage} and HELM \cite{liang2023holisticevaluationlanguagemodels}, designed to assess foundational LLM performance, are often presented as culturally neutral. 
However, recent analyses reveal demonstrate that performance on these benchmarks in fact requires considerable culturally contingent knowledge and assumptions. 
\citet{singh2025globalmmluunderstandingaddressing} found that 28\% of MMLU requires culturally-sensitive knowledge to answer correctly, demonstrating that accounting for cultural context can change system rankings. 

We argue that \textbf{cultural tasks should be expanded to include any task whose successful execution depends on cultural context, knowledge, norms, and user expectations.} 
In domains traditionally treated as ``value''-oriented, such as social bias or moral reasoning, culture-adapted benchmarks (\emph{e.g.}, \citet{jeong-etal-2022-kold},  \citet{jin2024kobbq}) have long embedded linguistic and socio-cultural conventions specific to their cultural contexts, reflecting the well-recognized influence of culture on these tasks. 
In contrast, tasks that are typically categorized as general capabilities, such as email writing or instruction-following, are often assumed to be culturally neutral and are evaluated without regard to contextual norms. Yet these tasks can be highly culturally contingent. 
For example, in Korean professional communication, emails to hierarchical superiors often begin with seasonal greetings or weather remarks, a practice rarely reflected in English-centric benchmarks. Performance on such subtle forms of localization remains largely unevaluated. While existing adaptations in value-oriented domains represent important progress, they overlook the broader set of tasks that also require cultural awareness to be executed appropriately.

\subsection{Task selection reflects Western priorities}\label{sec:unknowns}
Cultural evaluation also embeds implicit biases in determining which tasks are considered relevant or valuable.
As \citet{hershcovich2022challenges} argue through the concept of ``Aboutness,'' cultural context shapes what is considered important. Yet, current benchmarks often treat tasks as culturally neutral, applying them uniformly without regard for differing communicative goals, linguistic norms, or practical needs.
In practice, NLP evaluations routinely prioritize tasks rooted in English-speaking, Western contexts---often by adapting existing English benchmarks and framing non-English efforts as merely closing a performance gap. This bias is reinforced when task selection is based on user interaction data \cite{bhatt-diaz-2024-extrinsic}, which overwhelmingly reflects usage patterns in the U.S. and other Western nations \cite{zhao2024wildchat}.

This narrow framing has significant consequences. 
First, tasks meaningful primarily in Western contexts are often overrepresented.
For example, sentiment analysis of beer reviews is irrelevant where alcohol is prohibited~\cite{ji2020diversified}, and long-form news summarization may hold less value in cultures where news is already concise.
Second, and more critically, tasks crucial in other cultural contexts are underrepresented. English text refinement for non-native speakers---a vital need for millions globally---is one such example, often overlooked in mainstream evaluation.

Even the topics and categories underlying evaluation reflect Western assumptions.
For example, ostensibly universal notions like fairness and harm have largely been operationalized through Western categories, such as skin tone or race, leaving harms rooted in non-Western cultural contexts effectively unmeasured~\cite{qadri2025case,dammu-etal-2024-uncultured}.\footnote{For example, the GPT-4 Technical Report’s system card frames safety challenges primarily around Western-centric categories (\textit{e.g.}, race and gender).}
Yet, research shows that in practice, users from different cultural contexts engage with LLMs around very different concerns. \citet{tamkin2024clio} demonstrate that non-English conversations more often center on issues like economics, social concerns, or culturally specific content (\textit{e.g.,}anime). Similarly, \citet{kirk2024prismalignmentdatasetparticipatory} find that identity factors such as race and region have predictive power on the kinds of topics users discuss with LLMs, even when conversation framing is controlled.

To address these biases in task selection, we must move beyond simply adapting Western benchmarks and instead \textbf{build evaluation methodologies that emerge from and reflect the authentic needs and priorities of diverse user communities. }
A practical next step is to co-design evaluation tasks with these communities, ensuring they reflect real-world priorities and cultural norms. 
More broadly, identifying what we might call the ``unknown unknowns''---culturally significant capabilities, interaction patterns, and potential concerns that remain invisible to outside researchers---is crucial to developing LLMs that serve the global population without reinforcing existing power imbalances.

\section{How to evaluate}\label{sec:how}
Having established \textit{what} to evaluate, we now address \textit{how} to evaluate these diverse desiderata. Sometimes, \textit{what} can be feasibly evaluated is constrained by limitations in the \textit{how}. 

A major challenge in large-scale cultural evaluation is ``values pluralism,'' the existence of diverse, sometimes fundamentally irreconcilable perspectives \cite{berlin1969four}. As datasets grow to encompass more diverse sub-groups, core differences in perspective can render the aggregation across samples less meaningfully representative of a coherent ``culture'' as a whole \cite{diazScalingLawsNot2024}. 
This pluralism shapes how we can define and measure cultural alignment.

\subsection{Definitions of ``good'' are culturally contingent}\label{sec:definegood}

The primary manifestation of values pluralism in evaluation is that \textbf{what constitutes ``good'' behavior or desirable performance in a language model is itself culturally contingent and inherently subjective}. LM evaluation often seeks to assess ``good'' outputs, but there is no objective ``good'' when preferences are diverse and deeply rooted in cultural contexts.

Consider, for example, what patterns in responses to opinion questions make them distinctly American? \citet{johnsonGhostMachineHas} discuss how a propensity of ChatGPT to frame discussions of gun control legislation around individual liberties is a predominantly American position. However, this stance is neither uniquely nor comprehensively American. There are considerable populations of Americans who prioritize public safety over individual liberties in this debate and vice-versa. 
This illustrates a fundamental limitation: relying on a single viewpoint to model cultural representativeness will inevitably exclude significant internal diversity.
A more robust, albeit expensive, approach would involve demonstrating a language model's ability to understand and articulate the range of diverse perspectives that exist \textit{within} and \textit{between} societies.

This challenge extends to defining concepts critical for evaluation. \citet{lee-etal-2024-exploring-cross} show that annotators across English-speaking countries (US, UK, Australia, Singapore, South Africa) disagree significantly on what constitutes hate speech. Given that even related cultural contexts cannot agree on such a critical concept, this raises fundamental questions about developing universal classifiers or metrics for hate speech to evaluate language models in culturally-embedded tasks. This suggests that a bespoke metric tuned to the preferences of each culture being tested might be necessary.

Furthermore, the interpretation and use of evaluation scales themselves are subject to cultural variation, directly impacting how ``goodness'' is expressed and measured. 
\cite{Lee2002CulturalDI} find that Chinese and Japanese raters prefer midpoint satisfaction scores, as opposed to Americans readily providing high scores.
Individual endorsement of individualism also leads to less midpoint bias on questions that are otherwise unrelated to culture \cite{chen1995response}, suggesting that these culturally contingent values directly impact the meaning of scales for ``non-cultural'' tasks.

This phenomenon of ``extreme response style'' \cite{Chun1974ExtremeRS} between different cultures impacts a variety of domains, including online product and helpfulness ratings between Europe and Asia \cite{Barbro01012020} and differences in hotel and restaurant reviews in the Middle East and Anglosphere \cite{Alanezi2022UnderstandingTI}.
Such culturally contingent values inevitably impact model performance as diverse human preference feedback, reflecting these varied response styles, is collected and used for training or evaluation.

Beyond explicit opinions and scale use, culturally variable preferences exist for more nuanced desiderata like writing styles. Western readers, for instance, often have a stronger preference for concise and linear writing over more dialectical writing styles sometimes favored in some East Asian countries \cite{Kaplan1966CULTURALTPA,Shahid2024ExaminingHC}. Even within the Anglosphere, variations in national cultures drive differences in online communication styles \cite{Oprea2020TheEO}. 
Unlike simpler lexical or semantic similarity metrics \cite{zhang2019bertscore}, more complex, qualitative desiderata such as naturalness, engagingness, and understandability \cite{Zhong2022TowardsAU}, or likeability and interestingness \cite{Liu2023XEvalGM}, are extremely culturally variable and difficult to transfer across languages. 
Transferring these complex desiderata across languages is particularly challenging, as researchers cannot readily build on work developed predominantly in English and Western, Educated, Industrialized, Rich, and Democratic (WEIRD) contexts. A naive transfer risks unfairly penalizing outputs that align well with expected local cultural norms but deviate from WEIRD ones.

Acknowledging this cultural contingency is not to suggest an uncritical acceptance of all cultural norms. 
Rather, it is a necessary step to move past the current state of cultural ignorance and avoid the ``perspectival homogenization''~\cite{fazelpour2025value} of models to a single dominant viewpoint. 

\subsection{Reference examples alone cannot express culture as practice}\label{reference}

Given that definitions of ``good'' are culturally contingent and subject to pluralistic interpretations, evaluation paradigms that heavily rely on static reference examples or aggregated demonstrative samples inevitably struggle to capture this complexity. Such methods often implicitly assume a singular or dominant notion of correctness or preference, which, as discussed above, is an oversimplification.

This challenge manifests itself even in the simplest domains and evaluation metrics, such as in multiple-choice evaluation.
For instance, \textit{value alignment} research, which aims to move beyond evaluating culture as mere trivia, often captures \textit{culture as perspective} using demonstrative examples of culturally variable preferences on personality, political, and opinion questions, typically through questionnaires.

For example, \citet{alkhamissi2024investigating} frame cultural alignment in language models as the distributional similarity of models' answers to national populations on surveys like the World Values Survey \cite{inglehart2000world}.
While such work also seeks to adapt model affinity using interventions like persona-based prompting \cite{liChatGPTDoesntTrust2024}, the reliance on multiple-choice opinion outputs is problematic.
These multiple-choice opinion outputs from language models are notoriously noisy; \citet{khanRandomnessNotRepresentation2025} show how variations of opinions along value scales vary just as much under semantically-irrelevant stylistic modifications of the prompt as they do under cultural conditioning. 
Further, even when models authentically represent a distinct cultural perspective in their outputs, these questionnaire-based methods may miss them.
This calls into question the fundamental construct validity of questionnaire-based evaluations \cite{o1998empirical,davis2023benchmarks}. 

Static sets of exemplars can be problematic with more sophisticated metrics, too. 
Rich, context-dependent trained metrics can vary in unpredictable and task-dependent ways, with system scores that are completely contradictory with the same metric across different tasks.
For example, \citet{lum2024bias} note how simple ``trick tests'' of gender bias are not only not predictive of performance within a real-world task---such as generating English learning lessons and writing bedtime stories---but scores on these unrelated real-world tasks cannot predict each other.
These limitations point to the need for alternative evaluation designs that foreground cultural variation directly, rather than relying solely on static exemplars or task-specific metrics.

\subsection{Standard metrics are improperly situated}

Beyond diversifying representative \textit{samples}, we also need diverse representative \textit{metrics}. Metrics can encode many desiderata in ways that samples alone cannot.
However, conventional measures like accuracy or F1 assume a single correct answer and thus penalize culturally valid variation. 
Comparing model outputs to fixed ``correct'' references can miss problematic defaults, blind spots, or subtle stereotypes. For example, \citet{myung2024blend} highlight models repeatedly defaulting to narrow cultural artifacts (\emph{e.g.}, ``Seblak'' in West Java queries), a phenomenon invisible to standard qualitative metrics. 
This calls for a shift from singular metrics to multi-dimensional ones. For example, \citet{qadri2025case} show that evaluating cultural representation requires moving beyond factual accuracy to assess richer categories like the \textit{missingness} of iconic elements or the \textit{coherence} of cultural symbols.
Developing metrics that capture such fine-grained, socially-grounded dimensions is essential for moving beyond a simple pass/fail judgment of cultural alignment.
Addressing this gap requires moving beyond incremental tweaks toward pluralistic and structural alternatives \cite{sorensen2024roadmap}.

\section{In what circumstances to evaluate}\label{sec:where}
The circumstances of an evaluation are not culturally neutral. Yet current practices often fail to account for the deep cultural contingency embedded in two fundamental dimensions: (a) the language in which evaluation is conducted and (b) the interactional context it assumes.

\subsection{Language use is culturally situated}
Language is not a neutral vehicle for universal meanings; it embeds and enacts culture. The same concept can be realized through different linguistic forms depending on context, shaped by social and cultural norms that affect both form and style. Yet current evaluations often overlook this~\cite{hovy-yang-2021-importance, hershcovich2022challenges}, treating language as a simple variable rather than a cultural site. This oversight manifests in two common evaluation paradigms. First, in explicit ``cultural tasks,'' language often serves as a flat proxy for a culture, with evaluations focusing on task-specific performance parity (\textit{e.g.}, scoring knowledge about that culture) \cite{myung2024blend, shafayat2024multi, jin2024better}. Second, in seemingly universal tasks evaluated in a multilingual setting, language is treated merely as a constraint. In both cases, the methodology is confined to measuring whether a model's task performance remains consistent across different languages, obscuring the rich cultural information encoded within linguistic choices themselves.

A direct consequence of this methodological oversight is the failure to evaluate whether models respect the social and cultural norms embedded in language. For example, Korean has a complex honorific system reflecting social hierarchies~\cite{brown2015honorifics}. Evaluation in such contexts must assess not only informational correctness but also whether responses adhere to culturally appropriate politeness and formality—considerations less prominent in languages like English. A model response like ``좋은 질문이야!'' (\textit{Good question!}) may be grammatically correct yet pragmatically awkward, reflecting English conversational norms rather than Korean interactional expectations. Such mismatches clearly indicate failures of cultural alignment, even if the task's primary goal (\textit{e.g.}, answering a question) is met. Current evaluations typically restrict consideration of linguistic nuances to tasks like translation, neglecting them in instruction-following or question-answering scenarios where task-specific metrics dominate.

Furthermore, using language as a proxy for culture is problematic because the mapping is not one-to-one \cite{pawar2024surveyculturalawarenesslanguage, lee2023hate}; a single language can be used across many cultures, and a single culture can encompass multiple languages. This ambiguity challenges us to cautiously interpret performance gaps, recognizing that they can stem from a model's lack of cultural competence, the linguistic properties of the language itself, or an inseparable combination of both. For instance, \citet{saxon2023multilingual} demonstrate that performance disparities by language exist even on ostensibly non-cultural tasks such as common concept image generation. This shows that language itself is a powerful variable, making it difficult to isolate ``cultural knowledge'' as the sole factor behind performance differences in multilingual evaluations.

This critique does not dismiss the importance of performance parity across languages, which remains a crucial goal for multicultural equity. Rather, we argue that our approach to achieving it must be fundamentally expanded. A more robust evaluation framework would therefore address two distinct but related goals. First, it must move beyond informational correctness to assess pragmatic and cultural appropriateness, judging whether an utterance respects the social norms and communicative styles embedded in a language. Second, it must enrich the concept of ``parity'' itself, moving beyond task-specific metrics to include qualitative consistency. This involves using meta-metrics to track whether the granularity, amount, and quality of information remain stable across different linguistic contexts \cite{shafayat2024multi}. Together, these two advancements would shift evaluation from merely verifying if a model works in a language to assessing how well it communicates within that language’s cultural context.

\setlength{\textfloatsep}{10pt} 
\begin{figure}[t]
\centering
\begin{tcolorbox}[
  enhanced,
  boxrule=0.4pt,
  fonttitle=\bfseries,
  title=User reaction to ChatGPT's informal Korean output,
]\small
When you speak informally to ChatGPT, it now replies informally too, haha.  \\
(...) \\
I used to think of ChatGPT as my assistant, but when it suddenly spoke informally, I felt a bit offended, lol.  
I guess now I need to start thinking of it as more of a friend.\footnote{Originally posted in Korean on a public online forum. Source: \url{https://www.clien.net/service/board/park/18463114}}
\end{tcolorbox}
\caption{A Korean user reflects on ChatGPT’s unexpected use of informal speech, noting a shift in their perceived social relationship with the model. This illustrates the importance of speech-level appropriateness in culturally sensitive language generation.}
\label{fig:korean-speech-level}
\end{figure}

\subsection{Interaction patterns should be evaluated}
\label{sec: interaction}
Since the introduction of LLMs, especially ChatGPT and other web-based agents, conversational interactions have rapidly become the ``default'' interaction style for human-LLM engagement. This shift towards conversational, general-purpose chatbot models has fundamentally altered the landscape of evaluation, necessitating a more nuanced understanding of how interaction patterns themselves are culturally situated. Therefore, to evaluate LLMs for cultural alignment, we need to consider environmental and cultural differences not only in isolated, decontextualized statements but also \textbf{in the dynamics of interaction.} However, current cultural NLP research largely overlooks these nuanced interactional dynamics.


Cultural dynamics profoundly shape these human-AI interactions. Users from different backgrounds vary in their input styles, such as prompt directness across high and low-context cultures~\cite{haoyue2024factors}. Misinterpreting these culturally-specific instruction cues can cause LLMs to misunderstand intent and reduce conversation quality~\cite{chaves2021should}, creating disadvantages, especially in multi-turn interactions. Concurrently, users hold culturally grounded expectations for the AI's behavior and role, including politeness---as seen with Korean users seeking workarounds to ensure models maintain formality Figure~\ref{fig:korean-speech-level}---and the desired relational nature of the interaction, with some East Asian users seeking more rapport than typically task-focused Western users~\cite{folk2025cultural, ge2024culture}. How LLM manages these interactional styles significantly impacts user satisfaction and perceived quality.

However, \textbf{the way these cultural interaction style differences affect model performance is a major gap in current evaluation frameworks.}
While many studies report performance variations across languages~\cite{myung2024blend, shafayat2024multi, jin2024better}, the specific impact of culturally diverse interaction patterns remains largely unexplored. 
We lack comprehensive datasets representing diverse human-model interactions across cultures. Despite efforts like LMSYS~\cite{zheng2023lmsys}, Chatbot Arena~\cite{chiang2024chatbot}, and WildChat~\cite{zhao2024wildchat} collect ``in‑the‑wild'' interactions of users, these collections remain dominated by Western perspectives (53.7\% of WildChat logs are English queries, with 21.6\% of IP addresses from the United States and more than 40\% from Western countries).

This research gap is particularly concerning given that models demonstrate high sensitivity to prompt structure and phrasing \cite{dominguez2024questioning, zhu2023promptrobust, pezeshkpour-hruschka-2024-large}. 
Users whose natural communication patterns diverge from those dominant in training data may face consistent disadvantages in model performance and responsiveness, effectively experiencing a ``cultural prompt engineering tax'' that others do not.
This tax manifests at a fundamental level, with models often failing to reply consistently in the user's chosen language~\cite{marchisio-etal-2024-understanding}, forcing users to bear the extra cost of explicitly prompting ``Reply in Language X''.
Moreover, models consistently show degraded comprehension on code-switched text, a natural communication pattern in many multilingual or non-English speaking communities~\cite{mohamed2025lost}. 
Current approaches often place adaptation burdens on users rather than models (\textit{e.g.}, ``if the model isn't performing well, you're not prompting it correctly.'') This expectation, that users should conform to the model's preferred communication patterns rather than vice versa, demands critical rethinking.

Such cultural misalignments can have severe impacts, for example user alienation, trust erosion, and system abandonment by users from specific cultural backgrounds~\cite{adilazuarda2024towards}. 
This can create a self-reinforcing cycle: models become increasingly optimized for the cultural interaction patterns of those who continue to use them, while simultaneously becoming less accessible to others. 
Moreover, this dynamic risks what \citet{jones2025toward} describe as ``hegemonic interactional norms,'' where models trained predominantly on English-language data from Western contexts implicitly impose particular communication patterns on users from different backgrounds.


Therefore, evaluation frameworks must evolve to account for culturally diverse interaction styles. This means asking not only whether a model performs well overall, but whether it does so equitably across different cultural patterns of engagement. Addressing this requires: (1) collecting data on how users from diverse backgrounds naturally interact with LLMs—including turn-taking, request styles, and conversational repair; (2) analyzing how cultural expectations shape perceptions of response quality; and (3) developing interaction-focused metrics that assess a model's adaptability, identifying and mitigating performance disparities across interaction styles.

\section{Situated Researchers}\label{sec:situated}
Beyond the technical questions of what and how to evaluate cultural alignment lies a deeper set of socio-political questions concerning \textbf{who} performs this evaluation and \textbf{within what kind of research ecosystem}. The very practice of culturally-aligned evaluation is shaped by the positionality of researchers and the systemic biases embedded within the broader AI/ML community. 

The field's reliance on standardized benchmarks (\textit{e.g.}, GLUE~\cite{wang2018glue}, BigBench~\cite{srivastava2023beyond}, MMLU~\cite{hendrycks2021measuringmassivemultitasklanguage}) to characterize model capability and research value reinforces a subtle form of epistemic injustice.
Knowledge systems and problem formulations rooted in non-dominant contexts are often treated as peripheral---framed as ``extensions'' like ``benchmarks for X language''---rather than valued on their own terms. This reflects an implicit belief in the authority of dominant research centers to define legitimate knowledge, pressuring global researchers to conform by translating or adapting to English-centric benchmarks. In doing so, the current system risks marginalizing diverse epistemologies while treating English not merely as a lingua franca, but as the default arbiter of relevance and validity.

Researchers from non-Anglophone cultures face an implicit pressure: to gain visibility and legitimacy, their work must often first engage with English-centric tasks and benchmarks.
The pressure arises because work on English is routinely treated as a universal baseline rather than as research on one specific language—a tendency that the \emph{Bender Rule} directly critiques \citep{bender2011achieving}.  From an evaluation-research standpoint, this reality imposes an extra layer of labor: scholars must either (1) conduct parallel research (\textit{e.g.}, building two sets of dataset; one of their own the other English) or (2) first start with English to establish as a legitimate task and then move on to their own languages. 

However, language cannot be separated from culture. Just translating the problem at hand to English, or finding a superficially analogous English task often fails to address the phenomenon that originally motivated the research. For example, inferring social relationships from Korean dialogue is uniquely difficult due to the linguistic characteristics of Korean, such as frequent omission of the sentence subject, or Terms of Address that have unique social connotations, while it is less of a problem in other languages. 
In this sense, the global research ecosystem itself might actually be the primary bottleneck to developing genuinely \emph{culturally-aligned} language models.
It potentially hinders the development of research agendas truly grounded in diverse local contexts.

A meaningful shift in NLP evaluation thus requires more than new datasets or metrics. 
Evaluative choices---what to measure, how, and why---are shaped by positionalities, not objective truths. 
Focusing on simple trivia to characterize culture, while treating all ``non-cultural tasks'' as universal, hides bias behind a false veneer of objectivity.
We must (i) acknowledge our positionalities, (ii) seek out culturally-contingent aspects across \emph{all} evaluation domains, and (iii) embed local social, linguistic, and cultural expertise into dataset construction and protocol design.  
Only through this kind of multi-layered reflection can we hope to build NLP systems that are not only culturally meaningful but also globally inclusive.

\section{Implications and Future Directions}\label{sec:implication} 
\paragraph{Beyond decontextualized measures.}

While existing benchmarks serve as useful tools for comparing models’ general abilities (\autoref{sec:what}), they often fall short in evaluating how models perform in real-world, culturally situated contexts. 
Inspired by behavioral testing approaches like CheckList~\cite{ribeiro-etal-2020-beyond}, which systematically probe linguistic capabilities through targeted test cases, we propose extending existing “universal” benchmarks with explicit dimensions of cultural capability. By incorporating tests for ``cultural alignment failures''---such as how models handle culturally specific communication norms, contextually appropriate responses, or regionally relevant content. 

At the same time, as we discuss in \autoref{sec:how}, reference-based evaluation has fundamental limitations for cultural assessment: it cannot capture undesired default behaviors or accommodate the culturally contingent definitions of ``good'' we identified in \autoref{sec:definegood}. We need evaluation frameworks that accommodate multiple valid perspectives simultaneously rather than forcing consensus on a single metric. This requires benchmark designs that move beyond static references to holistically assess the acceptability and severity of cultural misalignments while systematically surfacing patterns of bias or insensitivity.

While employing LLM-as-a-judge systems is one promising method to evaluate such nuances without a single correct answer, as explored for cultural QA \cite{arora-etal-2025-calmqa}, we caution that this is not a silver bullet. Even when trained evaluators or LLM-as-a-judge systems are used, the cultural frame embedded in the judge itself risks re-inscribing a dominant perspective, creating epistemic circularity.


\paragraph{Discovering ``Unknown Unknowns''}
Human-centric evaluation of LLMs often masks LLMs’ unique, non-human errors; likewise, researchers evaluating cultures as outsiders risk overlooking problems they do not know exist. 
As discussed in \autoref{sec:what}, surfacing these ``unknown unknowns''---culturally meaningful tasks, interaction behaviors, and preferences currently invisible to us---is a major evaluation challenge.This includes both task-level gaps and interaction-level misalignments where cultural communication patterns affect model performance across user communities (\autoref{sec:where}).

To uncover these gaps, we need richer data on real user interactions, especially from underrepresented cultures. While resources like WildChat~\cite{zhao2024wildchat}, LMSys~\cite{zheng2023lmsys}, and Anthropic's Clio project~\cite{tamkin2024clio} provide useful insights, current datasets remain limited in cultural coverage and openness.

Addressing unknown unknowns also requires methodological support. Collaborating with HCI researchers can help; for example, interactive systems have been developed to visualize data gaps and guide human-in-the-loop data collection~\cite{10.1145/3706598.3713491}. 
The field also needs empirical studies on why researchers overlook these gaps and what interventions can help build more culturally robust evaluation practices~\autoref{sec:situated}.

\paragraph{Toward Stakeholder-Centered Evaluation Design.} 

To support more culturally responsive evaluation practices, we must identify and center those most directly impacted by LLMs (\autoref{sec:narrow_def}). 
Following \citet{smith2024recommend}, evaluation should involve stakeholders who can define appropriate behavior in context.

HCI research highlights the need for culturally sustaining practices that foreground community voices from the start~\cite{Anderson-Coto_Salazar_Lopez_Sedas_Campos_Bustamante_Ahn_2024}. Value-sensitive and participatory design approaches further warn against universalist assumptions,  emphasizing that evaluation standards must be situated in specific cultural contexts~\cite{friedman1996value, 10.1145/2207676.2208560}.

Recent research at the intersection of NLP and HCI shows that engaging stakeholders can surface overlooked dimensions of cultural representation~\cite{qadri2025case}.
We advocate for frameworks where stakeholders help define tasks, criteria, and evaluation standards through collaborative processes, moving beyond simply diversifying annotators. This participatory approach better aligns NLP evaluation with the real needs and values of affected communities.



\section{Conclusion}
We have argued that evaluation in language technology is never culturally neutral, and that every choice—explicit or implicit—carries cultural consequences. Our analysis shows that conventional evaluation practices, from task and metric selection to benchmarking standards, often obscure or marginalize diverse cultural realities. To move beyond these limitations, we advocate for culturally intentional evaluation: an approach that makes cultural context visible, explicit, and central at every stage of the evaluation pipeline. By centering positionality, engaging with affected communities, and embracing context-sensitive, “thick” evaluation practices, the NLP community can develop more equitable, representative, and impactful language technologies. We hope this work catalyzes further reflection and action, inviting researchers to critically reexamine and reimagine the cultural assumptions embedded in their evaluation practices, and to co-create more inclusive and responsive models for the world’s linguistic diversity.

\section*{Limitations}
While we advocate for a theory-driven and culturally intentional approach to evaluation in NLP, several limitations should be noted. 
First, this paper does not aim to be an exhaustive survey of all work in evaluations of cultural alignment or related fields. Readers seeking comprehensive overviews may refer to recent surveys such as \citet{pawar2024surveyculturalawarenesslanguage} or \citet{liu2024culturally}.
Additionally, our primary focus is on the evaluation of LLMs, which means that broader issues in language technology and culture are not discussed in detail.

As a position paper, our aim is to provoke discussion and outline future research directions, rather than to offer comprehensive solutions or empirical evaluations. We encourage further work that operationalizes these principles in a broader range of cultural, linguistic, and technological settings.

\section*{Acknowledgments}
This work was supported by Institute of Information \& communications Technology Planning \& Evaluation(IITP) grant funded by the Korea government(MSIT) (No. RS-2024-00509258 and No. RS-2024-00469482, Global AI Frontier Lab)

\section*{Contribution}
\textbf{Juhyun Oh} initiated and led the project, structured and wrote the manuscript, and contributed to the position and all sections. \\
\textbf{Inha Cha} initiated the project, contributed to the overall conceptualization, integrated STS and HCI perspectives into the implications, drafted the initial introduction, and refined all sections of the manuscript.\\
\textbf{Michael Saxon} contributed to the position, wrote initial drafts for Section 3, and contributed to revisions in other sections. \\
\textbf{Hyunseung Lim} contributed to the initial drafts for Section 4, and provided feedback on other sections.\\
\textbf{Shaily Bhatt} contributed to the overall conceptualization, wrote first drafts of section 2, and provided feedback on other sections. \\
\textbf{Alice Oh} advised the project and helped revise the manuscript.

\bibliography{custom}

\appendix
\section{Use of AI Assistant}
We used ChatGPT web assistant (ChatGPT Pro)\thinspace\footnote{\url{https://chatgpt.com/}} to refine the writing of the manuscript.



\end{document}